%% file: main.tex
\pdfoutput=1

\documentclass[11pt]{article}
\usepackage{arabtex}

\usepackage[]{acl}
\usepackage{times}
\usepackage{latexsym}
\usepackage[T1]{fontenc}
\usepackage[utf8]{inputenc}
\usepackage{microtype}
\usepackage{inconsolata}
\usepackage{graphicx}
\usepackage{amsmath}
\usepackage{graphicx}
\usepackage{multirow}
\usepackage{arabtex}
\usepackage{utf8}
\setcode{utf8} 
\usepackage{booktabs} 
\usepackage{longtable}
\usepackage{relsize}   
\usepackage[inline]{enumitem}
\usepackage{graphicx}

\usepackage{makecell}
\usepackage{xspace}
\usepackage{tcolorbox}

\newcommand{\toloka}{\texttt{Toloka}\xspace}

\newcommand{\jeem}{\texttt{JEEM}\xspace}

\newcommand{\jordanflag}{\includegraphics[width=5mm]{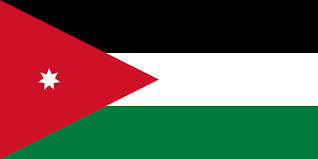} \xspace}
\newcommand{\uaeflag}{\includegraphics[width=5mm]{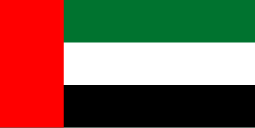} \xspace }
\newcommand{\egyptflag}{\includegraphics[width=5mm]{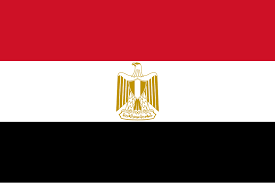} \xspace}
\newcommand{\moroccoflag}{\includegraphics[width=5mm]{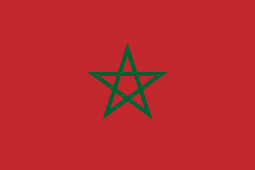} \xspace}

\usepackage[textsize=tiny]{todonotes}
\title{JEEM: Vision-Language Understanding in Four Arabic Dialects }

\author{
    \begin{minipage}[t]{\textwidth}
        \centering
        \normalfont
  Karima Kadaoui$^{1}$\footnotemark[1] \,
  Hanin Atwany$^{1}$\thanks{Equal contribution. \\Correspondence: \texttt{\href{mailto:karima.kadaoui@mbzuai.ac.ae}{karima.kadaoui@mbzuai.ac.ae}}} \,
  Hamdan Al-Ali$^{1}$\footnotemark[1] \,\\
  Abdelrahman Mohamed$^{1}$ \,
  Ali Mekky$^{1}$ \, 
  Sergei Tilga$^{2}$ \,
  Natalia Fedorova$^{2}$ \ \\
 Ekaterina Artemova$^{2}$ \,
  Hanan Aldarmaki$^{1}$\,
  Yova Kementchedjhieva$^{1}$\\ [1em]
        {
            \textsuperscript{1} MBZUAI
            \textsuperscript{2} Toloka AI
        }
    \end{minipage}
}

\begin{document}
\maketitle

\begin{abstract}

We introduce \jeem, a benchmark designed to evaluate Vision-Language Models (VLMs) on visual understanding across four Arabic-speaking countries: \textbf{J}ordan, The \textbf{E}mirates, \textbf{E}gypt, and \textbf{M}orocco\footnote{Our data is available at \url{https://huggingface.co/datasets/toloka/JEEM}.}. \jeem includes the tasks of image captioning and visual question answering, and features culturally rich and regionally diverse content. This dataset aims to assess the ability of VLMs to generalize across dialects and accurately interpret cultural elements in visual contexts. In an evaluation of five 
prominent open-source Arabic VLMs and GPT-4V, we find that the Arabic VLMs consistently underperform, struggling with both visual understanding and dialect-specific generation. While GPT-4V ranks best in this comparison, the model's linguistic competence varies across dialects, and its visual understanding capabilities lag behind. This underscores the need for more inclusive models and the value of culturally-diverse evaluation paradigms. 

\end{abstract}

\section{Introduction}
\input{intro}

\section{Related Work}
\input{related_work}

\section{Dataset Construction}
\input{data_collection.tex}

\section{Data Analysis}
\input{data_analysis}

\section{Benchmarking VLMs}
\input{Models}

\subsection{Image Captioning}

\input{evaluation_captions}

\subsection{Question Answering}
\input{evaluation_vqa}

\section{Conclusion}
\input{conclusion}

\section*{Limitations}
\input{limitations}

\section*{Ethics Statement}
\input{ethics_statement}

\bibliography{custom}

\appendix

\onecolumn

\section{Annotation Statistics} \label{sec:app_survey}

\subsection{Writer Profiles} \label{sec:app_task_distribution}

\input{app_survey}

\subsection{Tasks Distribution} \label{sec:app_task_distribution}

\input{app_task_distribution}

\clearpage

\section{Annotation Guidelines} \label{sec:app_annotation_guidelines}
\input{app_annotation_guidelines}

\section{Prompts}
\label{sec:eval_prompts}
\input{eval_prompts}

\section{Topic Categories}
\label{sec:app_topics_grouping}
\input{app_topics_grouping}

\end{document}

%% file: intro.tex
\begin{figure}[h!]
    \centering
    \includegraphics[width=.97\linewidth]{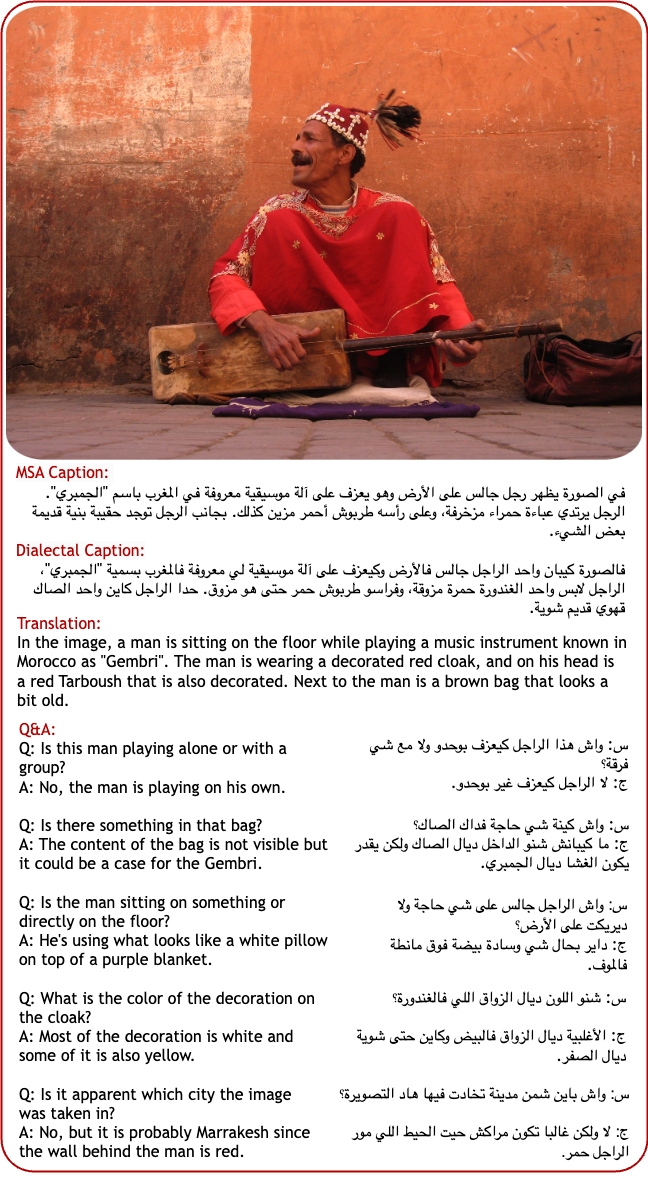}
    \caption{A sample from \jeem{} (Moroccan set). 
    }
    \label{fig:data-example}
    \vspace{-4mm}
\end{figure}

Vision-language models (VLMs) have recently achieved notable improvements in tasks such as image captioning (IC) and visual question answering (VQA), benefiting from large multimodal training datasets and parameter scaling \cite{grattafiori2024llama3herdmodels, beyer2024paligemma, openai2024gpt4technicalreport}. However, these models often struggle to generalize across culturally diverse and dialect-rich environments due to the over-representation of specific geographic regions \cite{de2019does,gustafson2023exploring} and standardized language varieties in their training datasets \cite{pouget2024no}. Similarly, existing evaluation datasets predominantly feature Western-centric images and English text \cite{liu-etal-2021-visually, wang2024cvluenewbenchmarkdataset}, while their non-English counterparts are often derived from the former, either through translation or relabeling of the same images \cite{changpinyo2023maxm}. This results in biased evaluation, which conceals the suboptimal performance of VLMs in geographically and dialectally diverse  settings \cite{bhatia2024local}. 

Recognizing this gap, recent work has focused on the creation of culturally diverse multilingual VQA benchmarks, incorporating images and questions from various countries and languages \cite[\textit{inter alia}]{liu-etal-2021-visually, pfeiffer2022xgqa, changpinyo2023maxm}. 
Among these, Arabic is rarely included, and when it is, it appears either in its standardized form (Modern Standard Arabic) \cite{tang2024mtvqa} or a single dialect, such as Egyptian \cite{romero2024cvqa}. This approach overlooks the cultural and dialectal diversity found among the $\sim$400 million speakers of this language.

Arabic is an official language in 25 countries across North Africa and the Middle East. Despite the shared language, each country has a different history, geography, and consequently culture.
These differences manifest in the objects, locations, and activities that visually characterize each region, as well as the lexical terms and implicit meanings associated with them.
For example, the traditional clothing item in the Gulf, the `kandura' (a long white robe worn by men) differs subtly from the `djellaba' worn in Upper Egypt, each reflecting regional identity and invoking different societal norms. On a linguistic level, differences are found not only in terms of lexicon, but also in phonetics and syntax, sometimes making mutual intelligibility challenging even among native Arabic speakers.

To address the challenges posed by the cultural and dialectal diversity of Arabic, we introduce \jeem, a benchmark dataset spanning one representative dialect from each dialectal region \cite{book-habash}: \textbf{J}ordanian (Leventine), \textbf{E}gyptian, \textbf{E}mirati (Khaleeji), and \textbf{M}oroccan (Maghrebi). \jeem comprises two core tasks: image captioning and visual question answering. These tasks enable the evaluation of VLMs in terms of their ability to recognize and appropriately reason about cultural elements, such as traditional clothing, local artifacts, and social settings, while utilizing  dialectal language. 

We benchmark five VLMs on JEEM and measure performance in terms of standard count-based metrics, GPT4-based evaluation, and human evaluation. This comprehensive  evaluation protocol allows us to reliably compare the five VLMs, identifying performance gaps in all models, including the top-ranking one, GPT-4o. 
We also evaluate the automatic metrics against human judgments and provide recommendations 
for future evaluations on JEEM and other Arabic benchmark datasets.

%% file: related_work.tex
\subsection{Why Culture Matters}

Prior studies reported performance disparities across cultures on machine learning tasks such as object recognition \cite{de2019does,gustafson2023exploring}, geolocalization \cite{pouget2024no}, mutlimodal retrieval \cite{kadar-etal-2018-lessons, buettner-kovashka-2024-quantifying} and visual question-answering \cite{romero2024cvqa}. 
These disparities are commonly
attributed to biases in the data on which models are trained, which tends to over-represent high-income geographic regions, and in particular Western ones \cite{de2019does,gustafson2023exploring, pouget2024no}.

People from different cultures use different objects, have different traditions, and occupy different physical environments, resulting in different visual experiences and associations. They also show perceptual differences with respect to specificity and importance \cite{nisbett2013culture}. Culture determines whether a general or a more specific term will be used to refer to an object, how the importance of background objects will be ranked with respect to foreground objects, and what objects will be mentioned in a caption or omitted \cite{nisbett2013culture, buettner-kovashka-2024-quantifying}. 

\subsection{Vision-Language Resources in Arabic}

\paragraph{Image Captioning}
Early work on Arabic IC \cite{Jindal_2017, mualla2018development} relied on the machine translation of existing datasets (primarily MSCOCO \cite{mscoco} and Flickr8K \cite{flickr8k}), sometimes including human validation \cite{eljundi2020resources} or human translation for a subset of the data \cite{Al-muzaini2018}.
AraCOCO \cite{mohamed-etal-2023-violet}  features 500 images from the MS COCO test set, captioned by Arabic speakers. 
While the annotators often mentioned details that did not appear in the original English caption, attesting to the difference in cultural perspectives, the captioned images were not sourced locally from the Arab-speaking world. Moreover, these captions are in MSA and follow the same short simplistic format found in COCO, lacking in dialectal and cultural understanding.

\paragraph{Visual Question-Answering}
While several multilingual datasets focus on culturally relevant VQA, many exclude Arabic \cite{gao2015you, gupta2020unified, pfeiffer2022xgqa, changpinyo2023maxm}. 

VAQA \cite{kamel2023vaqa} relabels MS COCO images in MSA, again limiting the cultural relevance and dialectal coverage of the data. 
Some works in text-only QA address this issue through manual cultural alignment \cite{alyafeai2024cidar} or sourcing data from Arab countries directly \cite{koto-etal-2024-arabicmmlu}, but lack the visual component. Others incorporate images but are limited to a single Arabic variety \cite{tang2024mtvqa, romero2024cvqa}, or rely on synthetic questions that are not always visually grounded \cite{alwajih-etal-2024-peacock}.

%% file: data_collection.tex
\jeem{} consists of images originating from four Arabic-speaking countries covering four distinct dialectal regions: Jordan (Levantine), Emirates (Gulf), Egypt (Egyptian), and Morocco (Maghrebi). Each image is annotated by native speakers of the target dialect with image captions in both MSA and dialect, and question-answer pairs in dialect.  

\paragraph{Team Organization and Recruitment}
The annotation process was led by four native speakers of the target dialects, each with a background in computational linguistics or natural language processing, hereafter referred to as team leaders.

The annotator recruitment process began with a free qualification task designed to identify annotators who met the following criteria:
\begin{enumerate*}[label=\itshape\roman*\upshape)]
\item had relevant professional experience; 
\item were native speakers of the target dialects;
\item  could produce high-quality image captions.
\end{enumerate*} As part of the qualification task, candidates wrote a caption for one image in both the target dialect and MSA. Each submission was carefully reviewed by a team leader. The candidates who performed best in terms of fluency and relevance were subsequently invited to join the project. This process led to the recruitment of 10, 8, 10, and  9 annotators for Jordan, the Emirates, Egypt, and Morocco, respectively. Their sociodemographic statistics, collected through a voluntary survey, can be found in Appendix~\ref{sec:app_survey}. 

\paragraph{Annotation setup}
The data collection process is based on how a visually impaired user might interact with a smart assistant: given an image with which the user wishes to engage (Step 1), the smart assistant would offer an initial description of the image (Step 2); At this point, the user might ask clarifying questions and inquire about further details (Step 3), to which the assistant would provide an answer (Step 4). We do not claim this procedure to accurately represent the experience and needs of visually impaired users, but it serves as a useful framework for guiding annotators on how to engage with the task, and for collecting natural questions born out of a genuine information scarcity. The process is visualized in Figure~\ref{fig:annot-pipeline} in Appendix \ref{sec:app_annotation_guidelines}. 

\begin{figure}[t]
    \centering
    \includegraphics[width=\linewidth]{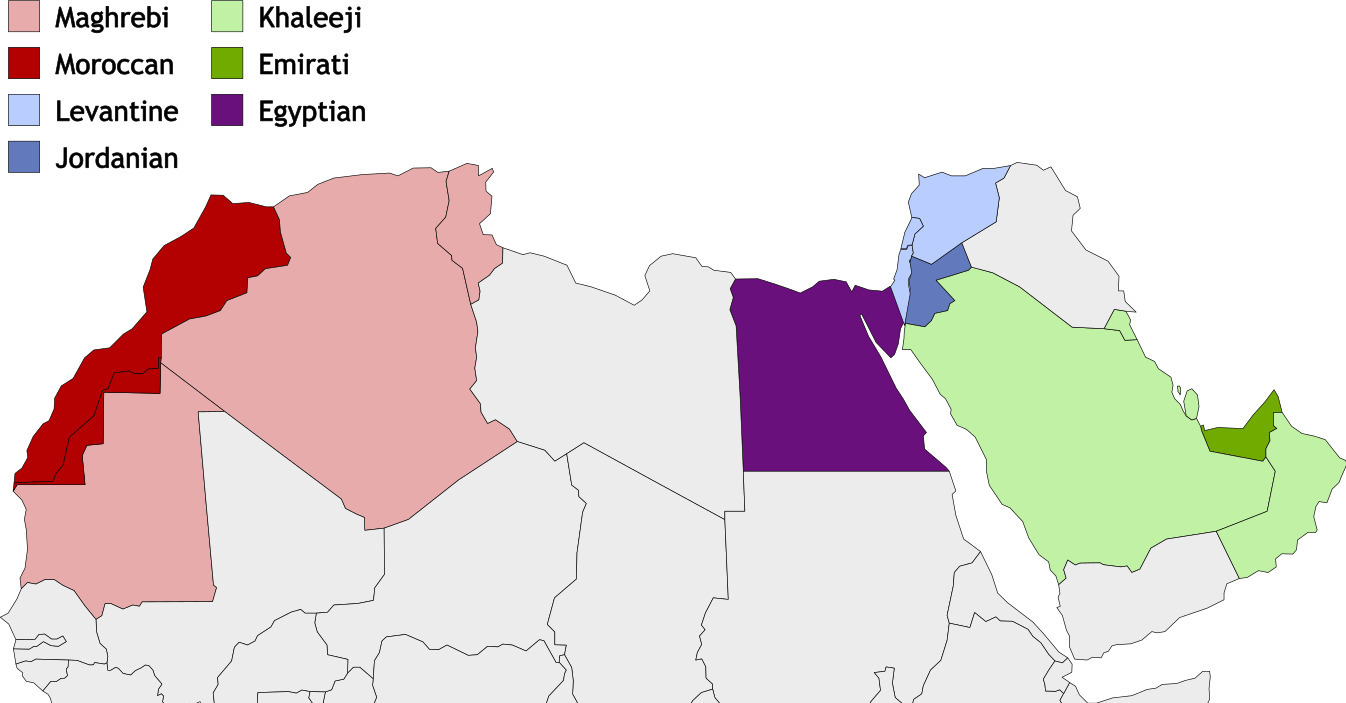}
    \caption{Dialectal coverage of \jeem. The country-level dialects used are shown in dark colors along with their respective region-level dialects in lighter color. The regional classification follows the work of \citet{book-habash}.}
    \label{fig:enter-label}
\end{figure}

\paragraph{Step 1: Image Collection} The objective of this step is to gather diverse, publicly available images that represent typical daily life in the target regions.  To this end, we collected images from three sources:
\begin{enumerate*}[label=\itshape\roman*\upshape)] 
\item Wikimedia archive, where images were sampled from categories under the tag \url{Category:<country>_by_topic} (all subject to a Creative Commons license).
\item Flickr archive under a Creative Commons license: the images were retrieved using tags such as country names, city names, and names of important places.
\item Personal archive: coauthors of this paper and team leaders contributed images from their personal collections that show typical scenes of daily life in their region of origin. 
They also reviewed and filtered all images sourced from Wikipedia and Flickr to ensure appropriate and informative selection. 
\end{enumerate*} 

Refer to \autoref{tab:Dialects} for final image counts in \jeem{}. 

\paragraph{Step 2: Image Captioning} The task is to write a description of the given image in both  Modern Standard Arabic (MSA) and dialect. Annotators were instructed to write in their dialect first to encourage spontaneous writing. They were instructed to provide descriptions that are  detailed enough to convey the content to someone who cannot see the image, including details specific to their region. 

\paragraph{Step 3: Question Writing} The task is to write five questions in dialect based on the given image description (the image is not shown to the annotator). The questions should be independent of each other and aim at a better understanding of what is happening in the unseen image. To avoid repeated exposure to images, annotators assigned to write a caption for a particular image were not assigned to write questions for the same image.

\paragraph{Step 4: Question Answering} The task is to answer five questions in dialect, based on the corresponding image and captions. If it is not possible to answer a question (e.g., the image does not contain the necessary information), annotators were instructed to indicate that the image lacks sufficient information. Answers should be based on the image and a general understanding of its context.

Annotators in each dialect group were assigned to annotate images specific to their region. In addition, a set of 25 images per dialect was manually selected to form a \textit{shared} pool of culturally distinctive places, dishes and objects. These images were annotated by all four regional teams to enable the exploration of cross-cultural perspectives.

\paragraph{Task review} Each submitted task was reviewed by the respective team leader. Reviewers could reject a task and reassign it to another annotator, edit and accept the task, or accept it as is.
Additionally, reviewers were allowed to skip a task if the image or the writing appeared inappropriate or irrelevant. The team leaders collaborated closely with the annotators throughout the annotation project, providing suggestions for improvement and exchanging feedback in a group chat.

\paragraph{Dialect diversity}  Each annotator could complete a limited number of tasks per day to avoid having a small number of annotators dominating the annotations. See Appendix \ref{sec:app_task_distribution} for the distribution of tasks completed by each annotator. To ensure dialect diversity, annotators were encouraged to use the most natural language for their local area. If they encountered an unfamiliar word or phrase in writing from previous steps, they were instructed to ask in the group chat for clarification.

The annotation was carried out on the \toloka{} platform. The detailed annotation guidelines made available to the annotators can be found in \autoref{sec:app_annotation_guidelines}.

%% file: data_analysis.tex
\input{stats}

\begin{figure}
    \centering
    \includegraphics[width=\linewidth]{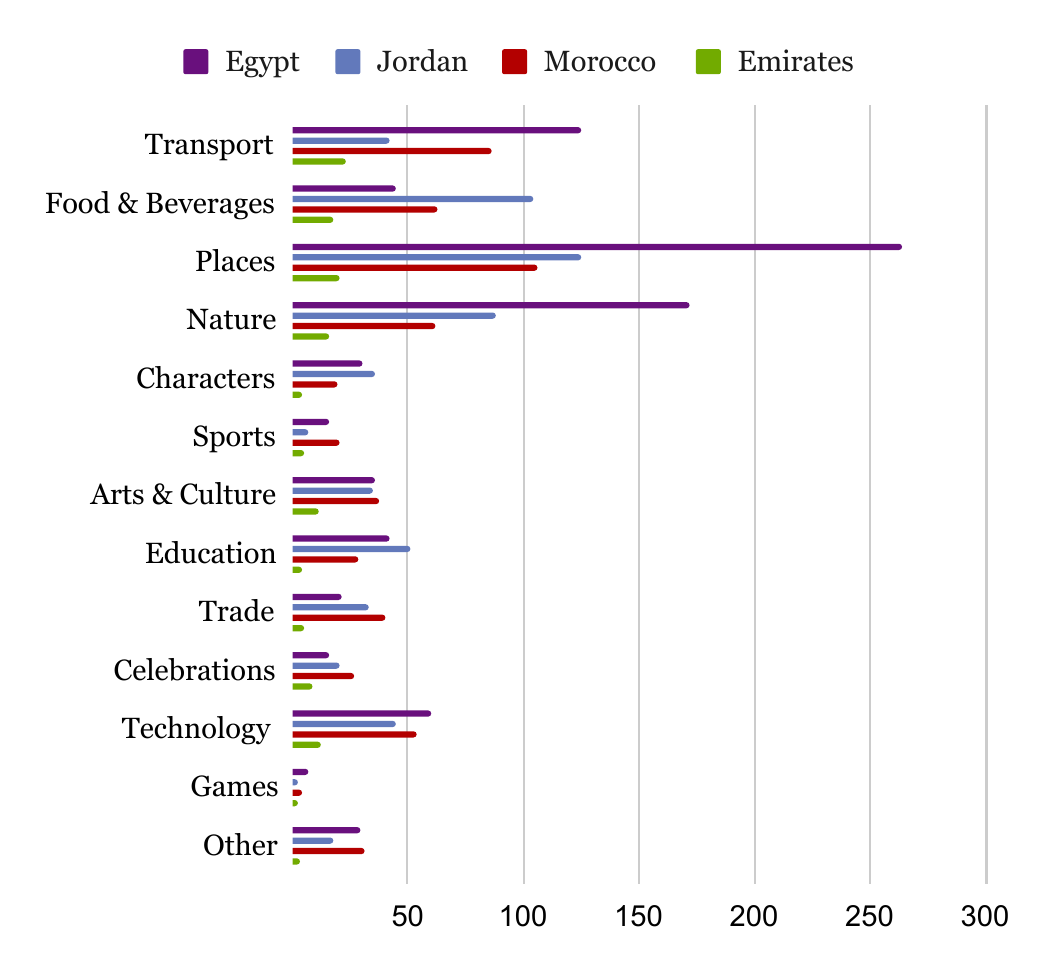}
    \caption{Topic distribution per dialect.}
    \label{fig:topic}
\end{figure}

\subsection{Data Distribution}
\jeem{} consists of 2,178 annotated images, distributed across the four dialects, as shown in Table~\ref{tab:Dialects}. The largest portion of images belongs to Egypt, followed by Jordan and Morocco, while the Emirates is represented in 6\% of the data. The table also reports statistics about the image captioning part of the dataset, including average caption length and the number of unique words used in each language variety. 

\paragraph{Image Topics} The images are organized into 13 thematic categories, including places, events, arts, nature, education, transport, food, trade, technology, characters, and games. These categories were identified after building the dataset, where we prompted GPT-4o mini \cite{openai2024gpt4technicalreport} in multiple rounds: first, to assign a topic to each MSA caption, and then to group the identified topics into the final categories (see the prompt in Appendix \ref{sec:eval_prompts}, Figure \ref{fig:topic_identification_prompt}). A detailed breakdown of subtopics within each category is provided in Table~\ref{tab:Topic Categories} in Appendix \ref{sec:app_topics_grouping}, while Figure~\ref{fig:topic} visualizes the topic distribution across dialects. Places emerge as the most common topic across images from all countries. However, the distribution of other prominent topics varies by region: nature was the second most frequent topic in Egypt, while food and beverages (F\&B) dominated in Jordan, and transport was prominent in both the Emirates and Morocco.

\paragraph{Image Captions} 
In written form, MSA and dialectal Arabic exhibit distinct variations in morphemes, sentence structure, and spelling conventions, which serve as indicators of dialectal influence in text \cite{keleg2023aldi}. This is evident in the variation of the average number of words used in captions across different dialects, as shown in Table~\ref{tab:Dialects}. Emiratis tend to use the fewest number of words in their captions, averaging 41 words per caption. In contrast, Egyptians write significantly longer captions, averaging 58 words per image.

\paragraph{Questions and Answers}
The total number of QA pairs across dialects is 10,890. In order to gain insight into the type of questions asked, we employed few-shot prompting of GPT-4o mini. The prompt defines four distinct question types (Descriptive, Quantitative, Categorical, and Yes/No) and, provides a detailed explanation of its defining characteristics with three examples in different dialects (see the prompt in the Appendix, Figure \ref{fig:question_type_prompt}). The distribution of question types across dialects is shown in Table \ref{tab:Question Type}, alongside some examples. The most prevalent type of questions is Descriptive, accounting for 45.92\% of the total, followed by Yes/No questions at 26.42\%, Categorical questions at 18.83\%, and Quantitative questions at 8.83\%.
\input{question_type}

\subsection{Cultural Aspects}
\begin{figure*}
    \centering
    \includegraphics[width=\linewidth]{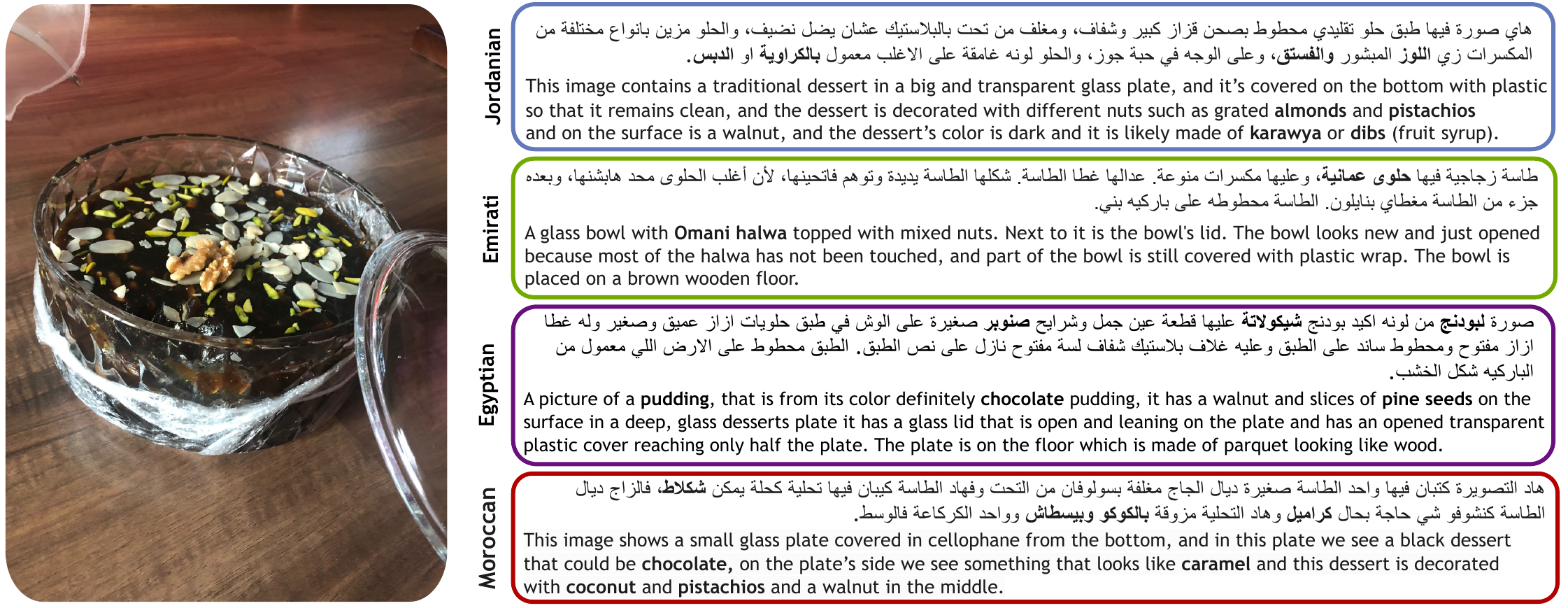}
    \caption{Image of a Omani Halwa (image sourced from the Emirati set) shared with annotators across all dialects. The Jordanian, Egyptian and Moroccan captions demonstrate an incorrect identification of the dessert and its components.}
    \label{fig:common}
\end{figure*}

We manually explored the shared pool of 100 images captioned in all four dialects to gain an understanding of how cultural perspective shapes perception.
One notable example is shown in Figure~\ref{fig:common}. It involves an image of Omani Halwa, a traditional Gulf dessert made from margarine, sugar, rose water, and semolina. As illustrated in the captions in the four dialects, only the Emirati annotator correctly identified it as Omani Halwa. In contrast, the Jordanian annotator misidentified it as Karawya
, a visually similar dessert with slight variations in texture and color, and both the Moroccan and Egyptian annotators mistakenly described it as a chocolate dessert, showcasing the diverse regional influences on object recognition.

%% file: stats.tex
\begin{table}[t]
\centering
\resizebox{\linewidth}{!}{
\begin{tabular}{l| c cc cc}
\toprule
\multirow{2}{*}{\text{\textbf{Country}}} & \textbf{Image} & \multicolumn{2}{c}{\underline{\textbf{Average Length}}} & \multicolumn{2}{c}{\underline{\textbf{Unique Words}}} \\

&\textbf{Count}& \textbf{DA} & \textbf{MSA} & \textbf{DA} & \textbf{MSA} \\
\midrule
\jordanflag Jordan & 606 & 46 & 52 & 8,933 & 9,751 \\

\uaeflag Emirates & 132 & 41 & 44 & 2,453 & 2,574 \\

\egyptflag  Egypt & 863 & 58 & 63 & 10,700 & 12,941 \\

\moroccoflag  Morocco & 577 & 52 & 52 & 7,822 & 8,161 \\
\bottomrule
\end{tabular}
}
\caption{Dataset statistics: image count, average caption length, number of unique words in the \jeem{} dataset.}
\label{tab:Dialects}
\end{table}

%% file: question_type.tex
\begin{table}[t]
\centering
\small
\begin{tabular}{l l}
\toprule
\vspace{2pt}
\textbf{Type:} Descriptive & \textbf{Percentage:} 45.92 \\
\textbf{Example:} &  \\
 \makecell[l]{What are the people\\on the roof wearing?} &  \makecell[r]{\scriptsize \<الناس اللي على السقف> \\ 
\scriptsize  \<لابسين إيه؟>} \\
\midrule
\textbf{Type:} Categorical &
\textbf{Percentage:} 18.83 \\
\textbf{Example:} & \\
\makecell[l]{Are these people in\\the kitchen men or women?} & \makecell[r]{\scriptsize \< هاد الناس اللي فكوزينة> \\
\scriptsize  \<واش رجال و لا عيالات؟>}\\
\midrule
\textbf{Type:} Quantitative &
\textbf{Percentage:} 8.83 \\
\textbf{Example} & \\
 \makecell[l]{How many boats can\\we see in the picture?} & \makecell[r]{\scriptsize \<كم طراد نقدر نشوفه في> \\
\scriptsize  \<  الصورة؟>}\\
\midrule
\textbf{Type:} Yes/No &
\textbf{Percentage:} 26.42 \\
\textbf{Example} & \\
 \makecell[l]{Does it look like they're\\cooking something on the stove?} & \makecell[r]{\scriptsize \<مبين انهم حاطين اشي> \\ 
\scriptsize  \<  عالنار؟>} \\
\bottomrule
\end{tabular}
\caption{Question type distribution across \jeem{}.}
\label{tab:Question Type}
\end{table}

%% file: Models.tex
We benchmark five recent Arabic-capable VLMs on \jeem{}: Maya \cite{alam2024maya}, PALO \cite{maaz2024palo}, Peacock \cite{alwajih-etal-2024-peacock}, AIN \cite{heakl2025ain}, and AyaV \cite{cohere2025aya}. All of these models were trained on Arabic data, sometimes authentic, but often translated from English, and showed good performance in Arabic captioning and VQA. 

For completeness, we also evaluate GPT-4 \cite{openai2024gpt4technicalreport} on \jeem, as it has been shown to achieve the best performance in Arabic tasks \cite{alyafeai2023taqyimevaluatingarabicnlp}.
We evaluate the models on three types of evaluations: traditional evaluation metrics, LLM-based evaluation, and human evaluation.

%% file: evaluation_captions.tex
\input{merged_evaluation}

\subsubsection{Traditional Evaluation Metrics}

We assess captioning performance using four standard metrics: CIDEr \cite{vedantam2015ciderconsensusbasedimagedescription}, ROUGE \cite{lin-2004-rouge}, BLEU \cite{papineni-etal-2002-bleu}, and BERTScore \cite{zhang2020bertscoreevaluatingtextgeneration}. For BERTScore, we use CamelBERT \cite{inoue-etal-2021-interplay}, which was trained on both MSA and dialectal Arabic. Following prior work \cite{zhang2020bertscoreevaluatingtextgeneration}, we report recall rather than F1 score. Given the morphological complexity and dialectal diversity of Arabic, we expect standard metrics to prove suboptimal. Therefore, we conduct  further evaluation using a large language model, namely GPT-4 \cite{mao-etal-2024-gpteval, tong2024gvevalversatilemetricevaluating}.

\input{gpt4o-output}

\subsubsection{GPT4-Based Evaluation}
We perform the evaluation according to four criteria, inspired by prior works \cite{liu-etal-2023-g}:
\begin{enumerate*}[label=\itshape\roman*\upshape)]
\item \textbf{Consistency} evaluates whether the caption matches what is shown in the image;
\item \textbf{Relevance} evaluates whether the caption describes the most important elements in the image;
\item \textbf{Fluency} evaluates how natural and fluent the text is;
\item \textbf{Dialect authenticity} evaluates whether the caption represents the target dialect.
\end{enumerate*} 
The evaluation was carried out using a five-point Likert scale, where 1 means failure to meet the criteria and 5, full compliance.

The evaluation is done in two setting: with access to both the input image and the reference, and with access to the reference alone \cite{tong2024gvevalversatilemetricevaluating}. This approach helps isolate the role of visual grounding in the evaluation process. The prompts used in each setting are detailed in \autoref{sec:eval_prompts}.

\subsubsection{Human Evaluation}

Since GPT-4 cannot be expected to have deep familiarity with and understanding of the cultural cues associated with each of the four regions covered by \jeem{}, we also conduct a human evaluation, using the same criteria as above. 
For each dialect group, we sampled a subset of 50 images and evaluated the models' predictions. To determine the ceiling of human performance, we also included the reference caption in this evaluation. Each caption was evaluated by a single annotator, blind to the origin of the caption. Refer to \autoref{sec:app_annotation_guidelines} for annotation guidelines.

Following prior work \cite{hessel2022clipscorereferencefreeevaluationmetric, wada2024polosmultimodalmetriclearning}, we compute Kendall’s Tau-C, a non-parametric statistic which indicates how well changes in one variable correlate with changes in another, and we begin the discussion below with a focus on this meta-metric, before moving on to discuss how models compare against each other based on the captioning metrics.

\subsubsection{Results}

\autoref{tab:merged_eval} reports model performance across the four subsets of \jeem{}. 

\paragraph{Metric quality}
Considering the traditional metrics first, we see that the range of the count-based metrics is lower than what is typically seen for English, for example. This is expected given the morphological richness of Arabic and its dialects. The range for BERTScore appears more promising. Yet, at the bottom of the table, we see that the correlation scores are low for all four metrics, and in fact lowest for BERTScore.\footnote{We measured the correlation for the F1 score derived from BERTScore and found an even lower, negative correlation.}

Moving to GPT4-based evaluations, we see a considerably higher correlation against human judgements in the image-and-reference setting (34.7), and an even higher correlation (39.1) with references alone. This contradicts the findings of \cite{tong2024gvevalversatilemetricevaluating}, who observed a higher correlation in the combined image-and-reference setting. We posit that the reference-only setting may reduce the bias of using GPT4 to evaluate GPT-4o. As we see, the scores considerably drop from the combined setting to the reference-only setting for all model predictions, but especially so for GPT4. 

\paragraph{Model comparison} Zooming in on this most reliable metric variant, GPT4 Eval with references, we see that among the five Arabic VLMs, AyaV achieves the best overall performance, especially in generating dialects compared to the other open source models. However, the competition is close with some models in other criteria. For example, AIN and Maya perform similarly in MSA, while Palo achieves comparable results in the Egyptian and Moroccan splits. Peacock, on the other hand, is generally weaker in terms of consistency, relevance, and fluency. Interestingly, however, it achieves the second-best dialectness score in Moroccan, with a significant margin over the closest competitor.
The objective which on average proves to be the easiest for all VLMs is fluency, while scores for all other criteria are low in comparison. This suggests that while the LM backbone is good at generating fluent MSA, the visual and dialectal aspects of the task is where the models can further improve.

\paragraph{Human evaluation} The scores indicate that GPT-4o is indeed a strong model for image captioning in Arabic, matching and surpassing human references on some criteria such as in the Jordanian and Egyptian splits. However, GPT-4o struggles considerably with the Emirati split of \jeem{}, in both linguistic and visual grounding terms. This points to the importance of having good regional coverage in the data used to train VLMs. Shifting the focus to open source models, their performance generally lags behind GPT-4o, consistent with GPT Eval results. Notably, AyaV still shows significant improvement in dialect generation over the other models.

%% file: merged_evaluation.tex
\begin{table*}[t]
\centering
\resizebox{\textwidth}{!}{
\begin{tabular}{c|l|cccc|cccc|cccc|cccc}
\toprule
 & \multirow{2}{*}{\textbf{Model}} & \multicolumn{4}{c|}{\textbf{Traditional Metrics}} & \multicolumn{4}{c|}{\textbf{GPT Eval (Image + Ref)*}} & \multicolumn{4}{c|}{\textbf{GPT Eval (Ref Only)*}} & \multicolumn{4}{c}{\textbf{Human Eval*}} \\
\cmidrule(lr){3-6}\cmidrule(lr){7-10}\cmidrule(lr){11-14} \cmidrule(lr){15-18}
 &  & \textbf{B} & \textbf{C} & \textbf{R} & \textbf{BSc} & \textbf{Con} & \textbf{Rel} & \textbf{Flu} & \textbf{DA} & \textbf{Con} & \textbf{Rel} & \textbf{Flu} & \textbf{DA} & \textbf{Con} & \textbf{Rel} & \textbf{Flu} & \textbf{DA} \\
\midrule
\multirow{6}{*}{\rotatebox{90}{ \textbf{MSA}}} 
  & AIN & 4.00 & 1.05 & 7.46 & 80.31 & 2.62 & 2.45 & 4.23 & - & 1.54 & 1.58 & 4.02 & - & - & - & - & - \\
  & AyaV    & 4.10 & 0.76 & 9.85 & 90.36 & 2.90 & 3.24 & 4.50 & - & 1.72 & 1.82 & 4.18 & - & - & - & - & - \\
  & Maya    & 4.25 & 1.79 & 9.47 & 90.35 & 2.32 & 2.52 & 4.00 & - & 1.54 & 1.60 & 3.80 & - & - & - & - & - \\
  & Palo    & 4.26 & 1.76 & 9.48 & \textbf{90.46} & 2.33 & 2.55 & 3.96 & - & 1.49 & 1.57 & 3.76 & - & - & - & - & - \\
    & Peacock & 2.08 & 1.51 & 7.18 & 84.24 & 1.73 & 1.90 & 3.82 & - & 1.16 & 1.21 & 3.07 & - & - & - & - & - \\
  & GPT-4o   & \textbf{5.87} & \textbf{7.27} & \textbf{10.61} & 90.35 & \textbf{3.50} & \textbf{3.62} & \textbf{4.62} & - & \textbf{1.90} & \textbf{2.22} & \textbf{4.26} & - & - & - & - & - \\
\midrule
\multirow{6}{*}{\rotatebox{90}{\jordanflag \textbf{JO}}} 
  & AIN    & 2.19 & 0.45 & 5.57 & 81.55 & 2.77 & 2.79 & 4.38 & 2.65 & 1.63 & 1.79 & 4.23 & 2.63 & 2.94 & 3.13 & 4.25 & 1.25 \\
  & AyaV    & 2.68 & 0.83 & 7.59 & 89.34 & 3.04 & 3.16 & 4.30 & 2.82 & 1.78 & 2.04 & 4.08 & 3.06 & 4.26 & 4.38 & 4.04 & 2.26 \\
  & Maya    & 1.91 & 0.49 & 6.44 & 90.16 & 2.50 & 2.60 & 4.12 & 2.60 & 1.54 & 1.66 & 3.94 & 2.72 & 3.08 & 3.42 & 4.04 & 2.22 \\
  & Palo    & 2.05 & 0.68 & 6.63 & \textbf{90.73} & 2.56 & 2.52 & 4.28 & 2.58 & 1.64 & 1.74 & 4.20 & 2.64 & 3.36 & 3.26 & 3.74 & 1.08 \\
    & Peacock & 1.55 & 1.88 & 5.91 & 83.57 & 2.32 & 2.36 & 3.56 & 2.52 & 1.42 & 1.56 & 3.48 & 2.60 & 2.32 & 2.28 & 3.46 & 1.32 \\
  & GPT-4o   & \textbf{5.23} & \textbf{6.91} & \textbf{9.66} & 90.72 & \textbf{3.84} & \textbf{4.00} & \textbf{4.72} & \textbf{3.32} & \textbf{2.36} & \textbf{2.82} & \textbf{4.58} & \textbf{3.70} & \textbf{4.84} & \textbf{4.88} & 4.66 & 3.56 \\
  & Human   & - & - & - & - & - & - & - & - & - & - & - & - & 4.74 & 4.76 & \textbf{4.78} & \textbf{4.82} \\
\midrule
\multirow{6}{*}{\rotatebox{90}{\uaeflag \textbf{AE}}} 
  & AIN    & 1.63 & 0.52 & 5.03 & 81.89 & 2.59 & 2.76 & 4.20 & 2.07 & 1.49 & 1.60 & 4.15 & 2.06 & 3.34 & 3.62 & 4.74 & 1.00 \\
  & AyaV    & 1.69 & 0.88 & 5.80 & \textbf{90.24} & 2.94 & 3.20 & 4.34 & 2.22 & 1.70 & 1.76 & 4.18 & 2.34 & 3.86 & 4.16 & 4.74 & 1.14 \\
  & Maya    & 1.55 & 0.36 & 5.98 & 89.43 & 2.44 & 2.42 & 4.20 & 2.04 & 1.52 & 1.58 & 4.16 & 2.06 & 1.84 & 2.20 & 4.56 & 2.00 \\
  & Palo & 1.50 & 0.29 & 5.75 & 89.43 & 2.53 & 2.71 & 4.16 & 1.88 & 1.57 & 1.69 & 4.12 & 1.90 & 2.38 & 2.78 & \textbf{5.00} & 1.00  \\
    & Peacock & 1.23 & 0.95 & 4.09 & 79.09 & 2.04 & 2.06 & 3.86 & 1.82 & 1.22 & 1.33 & 3.84 & 1.73 & 2.28 & 2.84 & 4.00 & 1.00 \\
  & GPT-4o   & \textbf{3.19} & \textbf{2.73} & \textbf{7.21} & 89.03 & \textbf{3.58} & \textbf{3.84} & \textbf{4.74} & \textbf{2.58} & \textbf{1.96} & \textbf{2.34} & \textbf{4.54} & \textbf{2.62} & 3.24 & 3.32 & 4.88 & 2.20 \\
  & Human   & - & - & - & - & - & - & - & - & - & - & - & - & \textbf{4.48} & \textbf{4.72} & 4.92 & \textbf{4.94} \\

\midrule
\multirow{6}{*}{\rotatebox{90}{\egyptflag \textbf{EG}}} 
  & AIN & 2.08 & 0.31 & 5.20 & 79.60 & 2.40 & 2.36 & 4.23 & 2.47 & 1.44 & 1.58 & 4.06 & 2.23 & 3.67 & 3.79 & 4.27 & 2.21 \\
  & AyaV    & 2.87 & 0.83 & 7.85 & 89.82 & 2.76 & 2.80 & 4.26 & 3.78 & 1.74 & 1.94 & 4.04 & 4.02 & 3.58 & 4.18 & 3.70 & 3.44 \\
  & Maya & 2.16 & 0.49 & 6.67 & \textbf{90.82} & 2.06 & 2.14 & 4.00 & 2.46 & 1.40 & 1.48 & 4.04 & 2.36 & 3.76 & 3.12 & 2.88 & 1.64 \\
  & Palo & 2.05 & 0.52 & 6.37 & 91.10 & 2.13 & 2.55 & 4.18 & 2.55 & 1.61 & 1.78 & 4.10 & 2.37 & 3.48 & 4.02 & 4.50 & 1.74 \\
    & Peacock & 0.86 & 0.54 & 4.50 & 81.88 & 2.04 & 1.98 & 3.92 & 2.32 & 1.28 & 1.34 & 3.78 & 2.24 & 3.14 & 2.52 & 3.40 & 1.66 \\
  & GPT-4o & \textbf{4.09} & \textbf{8.41} & \textbf{8.56} & 90.64 & \textbf{3.16} &\textbf{ 3.40} & \textbf{4.64} & \textbf{3.88} & \textbf{1.88} & \textbf{2.16} & \textbf{4.38} & \textbf{4.08} & 4.44 & \textbf{4.36} & 3.70 & 4.34 \\
  & Human & - & - & - & - & - & - & - & - & - & - & - & - & \textbf{4.46} & 4.16 & \textbf{4.62} & \textbf{4.90} \\
\midrule
\multirow{6}{*}{\rotatebox{90}{\moroccoflag  \textbf{MA}}} 
  & AIN & 1.34 & 0.58 & 3.40 & 81.37 & 2.54 & 2.75 & 4.40 & 1.69 & 1.44 & 1.50 & 4.06 & 1.27 & 4.00 & 3.46 & 4.35 & 1.00 \\
  & AyaV    & 2.21 & 0.53 & 6.55 & 88.33 & 2.64 & 2.98 & 3.98 & 3.56 & 1.78 & 1.92 & 3.92 & 3.94 & 3.72 & 3.64 & 3.28 & 3.04 \\
  & Maya & 1.06 & 0.37 & 3.79 & 88.85 & 2.26 & 2.54 & 4.10 & 1.72 & 1.48 & 1.56 & 3.98 & 1.32 & 3.09 & 3.28 & 4.01 & 2.32 \\
  & Palo & 1.06 & 0.46 & 3.76 & 89.47 & 2.66 & 2.78 & 4.20 & 1.74 & 1.60 & 1.72 & 4.18 & 1.28 & 4.14 & 3.80 & 4.70 & 1.00 \\
    & Peacock & 0.51 & 0.40 & 2.55 & 79.88 & 1.98 & 1.98 & 2.72 & 2.04 & 1.36 & 1.42 & 2.66 & 2.48 & 3.56 & 2.56 & 3.86 & 1.00 \\
  & GPT-4o & \textbf{4.73} & \textbf{6.70} & \textbf{9.00} & \textbf{89.98} & \textbf{3.48} & \textbf{3.50} & \textbf{4.70} & \textbf{4.12} & \textbf{2.04} & \textbf{2.48} & \textbf{4.38} & \textbf{4.28} & 4.64 & 4.59 & 4.55 & 4.38 \\
  & Human & - & - & - & - & - & - & - & - & - & - & - & - & \textbf{4.84} & \textbf{4.70} & \textbf{4.87} & \textbf{4.97} \\
  \midrule
  & $\tau_c$ & 23.9 & 24.5 & 18.5 & 13.2 & \multicolumn{4}{c|}{34.7} & \multicolumn{4}{c|}{39.1} & - & - & - & - \\
\bottomrule
\end{tabular}
}
\caption{Image Captioning. Combined evaluation results for different Arabic varieties. GPT-based evaluation results are split into two settings: Image + Reference Caption and Reference Caption Only. Human evaluation is also included where available. Automatic metrics are computed on the full \jeem{} dataset. The rest of the metrics are computed on a sample of the dataset. * Metrics measured on the same sample of 200 images. $\tau_c$: Kendal's Tau-C.}
\label{tab:merged_eval}
\end{table*}

%% file: gpt4o-output.tex
\begin{table}[]
    \centering
    \begin{tabular}{l}
    \toprule
         \makecell[c]{\textbf{\scriptsize Fig. \ref{fig:data-example}}} \\
         \makecell[r]{\scriptsize \<هذه صورة لحلوى شبيهة بالمغلي، مزينة باللوز والجوز ومكسرات أخرى.>\\ \scriptsize \<موجودة في وعاء زجاجي على طاولة خشبية. يبدو أن الحلوى مغطاة>\\ \scriptsize     \<  بورق شفاف للحفظ.>}\\
         \makecell[l]{\scriptsize This is an image of a sweet like \textbf{Meghli}, decorated with almonds and\\ \scriptsize    walnuts and other nuts. It is in a glass contained on a wooden table.\\ \scriptsize  It seems that the \textbf{sweet is covered} in cling  wrap for  conservation.}\\
         \midrule
         \makecell[c]{\textbf{\scriptsize Fig. \ref{fig:common}}}\\ \makecell[r]{\scriptsize \<هذا الراجل قاعد في الزنقة لابس لباس تقليدي أحمر، وعندو آلة موسيقية>\\ \scriptsize \<  غادية تكون الكنبري. كيبان بحال شي فنان شعبي كيقدم عرض ديالو.>\\ \scriptsize \<  الحيط اللي وراه ببرطقي، والجو العام كيعطي إحساس ديال المغرب العريق.>\\ } \\
        \makecell[l]{\scriptsize This man is sitting in the street wearing traditional red \textbf{clothing}, and he\\ \scriptsize  has a musical instrument that could be the Gembri. He looks like a folk\\ \scriptsize  artist showcasing his act. The wall behind him is \textbf{[unk]} and the general\\ \scriptsize   atmosphere gives a feeling of ancient Morocco.}\\
    \bottomrule
    \end{tabular}
    \caption{GPT4-o outputs on Figure \ref{fig:data-example} image in Emirati dialect (top), and Figure \ref{fig:common} image in Moroccan dialect (bottom). The model fails to provide a caption in the Emirati dialect, misidentifies Omani Halwa as Meghli (also known as Karawiya), and incorrectly states that the wrap covers the sweet itself. It has better fluency in Moroccan, but provides a vague description of cultural Moroccan clothing, and uses a word that appears dialectal but does not actually exist.}
    \label{tab:my_label}
\end{table}

%% file: evaluation_vqa.tex
\subsubsection{Evaluation Metrics}
We evaluate the VQA capabilities of the five models automatically using GPT4, using the same criteria as described above, with both the input image and reference answer. We expect that in questions of the descriptive type which are open-ended and allow for variability in valid answers, this would be a more suitable evaluation compared to traditional metrics.
We conduct GPT‑4–based evaluation to better assess how well each model interprets and integrates visual cues\footnote{The cost of GPT evaluation was \$29.125 for the questions and \$15.65 for the captions.}.  The prompts are detailed in \autoref{sec:eval_prompts}.
\input{gpt_evaluation_qa}

\subsubsection{Results}
The results  are presented in \autoref{tab:gpt_eval_qa}. Again, we see that the fluency criterion yields the highest scores across all models and dialects. However, the semantic criteria of consistency and relevance lag well behind, especially for the Arabic VLMs. This suggests that the model-generated answers do not successfully address the user question. The dialectness of the Arabic VLMs is similarly low across most dialects. Meanwhile, GPT-4o scores the best overall, despite some room for improvement on the semantic criteria, and reduced linguistic capabilities on the Emirati dialect. These findings align with the observations made for the image captioning task, and with the general motivation for the work, which is to expose the limitations of VLMs in grasping the semantics of culturally-loaded images across different Arabic-speaking regions and in expressing these semantics in the appropriate dialectal terms and syntactic structures. 




%% file: gpt_evaluation_qa.tex
\begin{table}[t]
\centering
\renewcommand{\arraystretch}{0.85} 
\scriptsize  

\small

\begin{tabular}{c|l|cccc}
\toprule
& Model  & \textbf{Con} & \textbf{Rel} & \textbf{Flu} & \textbf{DA} \\
\midrule
\multirow{5}{*}{\rotatebox{90}{\tiny\jordanflag \textbf{JO}}}  
& AIN & 2.41 & 2.55 & 4.11 & 3.04 \\
& AyaV & 2.76 & 2.96 & 4.22 & 2.55 \\
& Maya & 2.56 & 2.68 & 4.08 & 2.93 \\
& Palo & 2.39 & 2.48 & 4.08 & 2.89 \\
& Peacock & 1.40 & 1.41 & 2.61 & 2.31 \\
& GPT-4o & \textbf{3.56} & \textbf{3.70} & \textbf{4.67} & \textbf{4.26} \\
\midrule
\multirow{5}{*}{\rotatebox{90}{\tiny\uaeflag \textbf{AE}}}  
& AIN & 2.78 & 2.83 & 4.19 & 2.93 \\
& AyaV & 3.00 & 3.02 & 4.26 & 2.57 \\
& Maya & 2.61 & 2.72 & 4.17 & 2.95 \\
& Palo & 2.51 & 2.50 & 4.07 & 2.83 \\
& Peacock & 1.90 & 1.99 & 3.44 & 2.51 \\
& GPT-4o & \textbf{3.64} & \textbf{3.72} & \textbf{4.56} & \textbf{3.87} \\
\midrule
\multirow{5}{*}{\rotatebox{90}{\tiny\egyptflag \textbf{EG}}}  
& AIN & 2.18 & 2.26 & 3.72 & 2.92 \\
& AyaV & 2.63 & 2.72 & 4.10 & 2.66 \\
& Maya & 2.20 & 2.32 & 4.00 & 3.08 \\
& Palo & 2.07 & 2.09 & 3.90 & 3.06 \\
& Peacock & 1.44 & 1.47 & 3.05 & 2.41 \\
& GPT-4o & \textbf{3.26} & \textbf{3.36} & \textbf{4.58} & \textbf{4.54} \\
\midrule
\multirow{5}{*}{\rotatebox{90}{\tiny\moroccoflag \textbf{MA}}}  
& AIN & 2.22 & 2.36 & 3.68 & 2.37 \\
& AyaV & 2.60 & 2.78 & 4.06 & 2.10 \\
& Maya & 2.06 & 2.23 & 3.77 & 2.30 \\
& Palo & 2.00 & 2.14 & 3.76 & 2.21 \\
& Peacock & 1.43 & 1.47 & 2.77 & 2.24 \\
& GPT-4o & \textbf{3.52} & \textbf{3.71} & \textbf{4.58} & \textbf{4.56} \\
\bottomrule
\end{tabular}

\caption{GPT4-based evaluation of \textbf{question answering}, results for different Arabic varieties.}
\label{tab:gpt_eval_qa}
\end{table}

%% file: conclusion.tex
In this paper, we presented \jeem, a culturally-informed benchmark for VLMs that covers four diverse Arabic-speaking countries: \textbf{J}ordan, the \textbf{E}mirates, \textbf{E}gypt, and \textbf{M}orocco. By incorporating both image captioning and visual question answering, \jeem allows for a comprehensive evaluation of VLM generalization capabilities across different Arab cultural and dialectal contexts.
We also included an evaluation of four Arabic VLMs: Peacock, Maya, AIN, and Palo, in addition to the multilingual GPT-4o. We included both human and automatic evaluation results to ensure reliability, and attempted to measure performance across various dimensions. Our results indicate that current models, whether general or Arabic-specific, still struggle with dialectal and cultural understanding. GPT-4o achieved the highest scores on most metrics, but there is large room for improvement in semantic dimensions like relevance and consistency. In addition, models struggle more with low-resource dialects like Emirati compared to high-resource variants like MSA and Egyptian. 

%% file: limitations.tex
While \jeem provides a culturally diverse benchmark for VLMs, several limitations should be acknowledged.
First, the dataset focuses on only four Arabic dialects, leaving out many others and thus limiting a comprehensive and inclusive evaluation.
Second, although \jeem serves as a benchmark rather than a training dataset, its size remains relatively small compared to existing Western vision-language datasets. 

Additionally, automatic evaluation metrics such as CIDEr and BLEU may not fully capture the complexity of dialect-specific and culturally nuanced responses. While we incorporate human evaluation, it is conducted on only a subset of the data. Expanding human evaluation to a broader sample could provide a more comprehensive assessment of model performance.

%% file: ethics_statement.tex
\paragraph{Fair Job Conditions} Our team of writers is based in the United Arab Emirates, Jordan, Morocco, and Egypt. Their pay rates exceed the respective hourly minimum wages. Annotation and voluntary survey results are collected and stored anonymously. Writers are informed in advance about potentially sensitive or harmful content in the images, which may be related to topics such as politics, culture, and religion.

\paragraph{Licensing Information} The images are subject to the underlying licensing terms of Wikimedia Commons \footnote{\href{https://commons.wikimedia.org/wiki/Commons:Licensing/en}{\texttt{https://wikimedia.org/Licensing/}}} and Flickr
\footnote{\href{https://www.flickrhelp.com/hc/en-us/articles/4404070159636-Creative-Commons}{\texttt{https://flickrhelp.com/creativecommons/}}}. The image captions, questions, and answers are distributed under the MIT license\footnote{\url{https://opensource.org/license/mit}}.

%% file: app_survey.tex
\input{app_demo_survey}

%% file: app_demo_survey.tex
\begin{table}[htp!] 
    \centering
    \scriptsize
    \begin{tabular}{p{4cm} p{10cm}}
    \toprule 
    \textbf{Question} & \textbf{Response (\%)} \\
    \midrule
    \textbf{What gender do you identify as? } &
    Male: 45.8, Female: 54.2, Nonbinary/Other: 0 \\ 
    \midrule
    \textbf{What is your age? } &
    20-29: 50, 30-39: 29.2, 40-49: 20.8, 50+: 0 \\
    \midrule
    \textbf{What is your nationality? } & 
    Jordan: 37.5, Egypt: 29.2, Morocco: 20.8, UAE: 12.5 \\
    \midrule
    \textbf{What is your native language? } & Arabic: 95.8, Multiple incl. Arabic: 4.2 \\
    \midrule
    \textbf{What is your native dialect? } & Jordanian: 37.5, Egyptian: 29.2, Darija: 20.8, Emirati: 12.5 \\
    \midrule
    \textbf{Where did you grow up? (Nearest city)} & \makecell[l]{Jordan: Amman (33.3), Irbid (4.2); \\
    Morocco: Tetouan (8.4), Casablanca (8.4), Khenifra (4.2); \\
    Egypt: Cairo (8.4), Giza (4.2), Mansoura (4.2), Tanta (4.2), Damietta (4.2), Helwan (4.2); \\
    the Emirates: Al Ain (4.2), Abu Dhabi (4.2), Ajman (4.2)} \\
    \midrule
    \textbf{Highest level of education? } & High school: 4.2, Undergraduate: 41.7, Postgraduate: 29.2, Master's: 29.8, Doctorate: 4.2 \\
    \midrule
    \textbf{Years of work experience?} & 1-3: 37.5, 4-6: 12.5, 7-9: 16.7, 10-12: 16.7, 13-15: 4.2, 16+ years: 12.5 \\ 
    \midrule
    \textbf{What is your current employment status?} & Not working: 8.2, Self-employed: 25, Part-time: 33.3, Full-time: 33.3 \\

    \bottomrule
    \end{tabular}
    \caption{Results of the voluntary survey of 24 respondents.}
    \label{tab:writer_sociodem_survey}
\end{table}

%% file: app_task_distribution.tex
\begin{figure*}[h]

    \centering
    \includegraphics[width=0.99\linewidth]{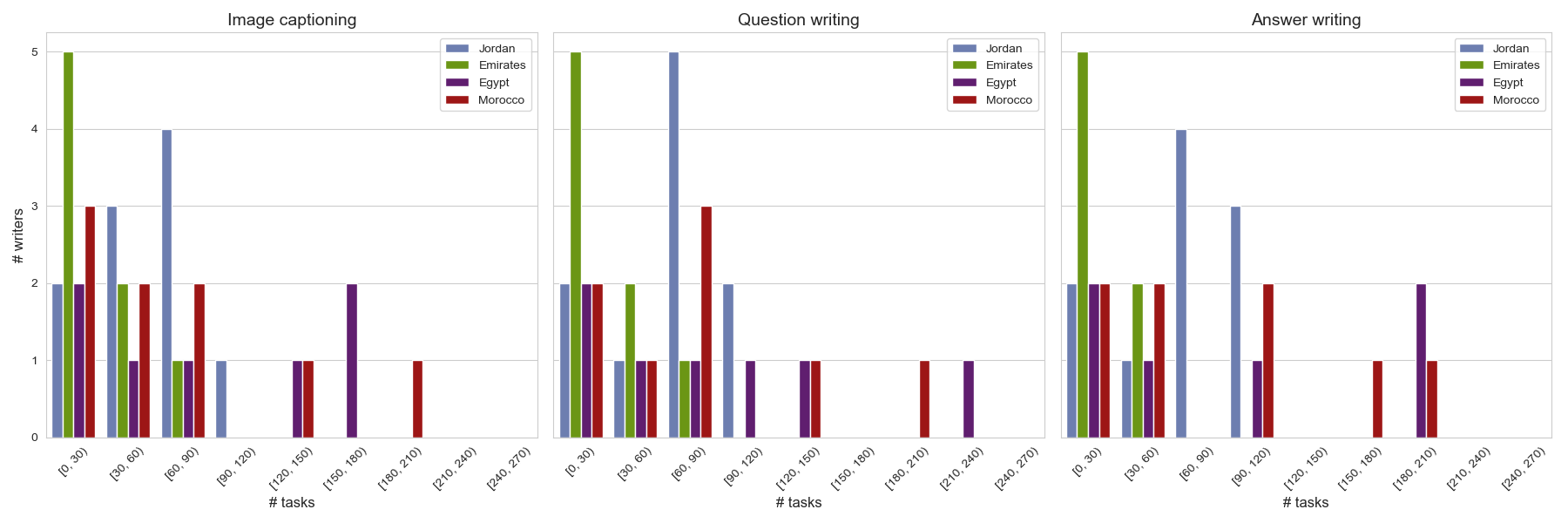}
    \caption{Distribution of annotators based on the number of tasks completed for three tasks: Image Captioning, Question Writing, and Answer Writing. Each bar represents the number of writers contributing within a given range, with colors indicating different dialects. Y-axis: number of unique writers. X-axis: the number of tasks grouped into intervals.}
    \label{fig:task_counts}
\end{figure*}

%% file: app_annotation_guidelines.tex
\begin{figure*}[htp!]
    \centering
    \includegraphics[width=\linewidth]{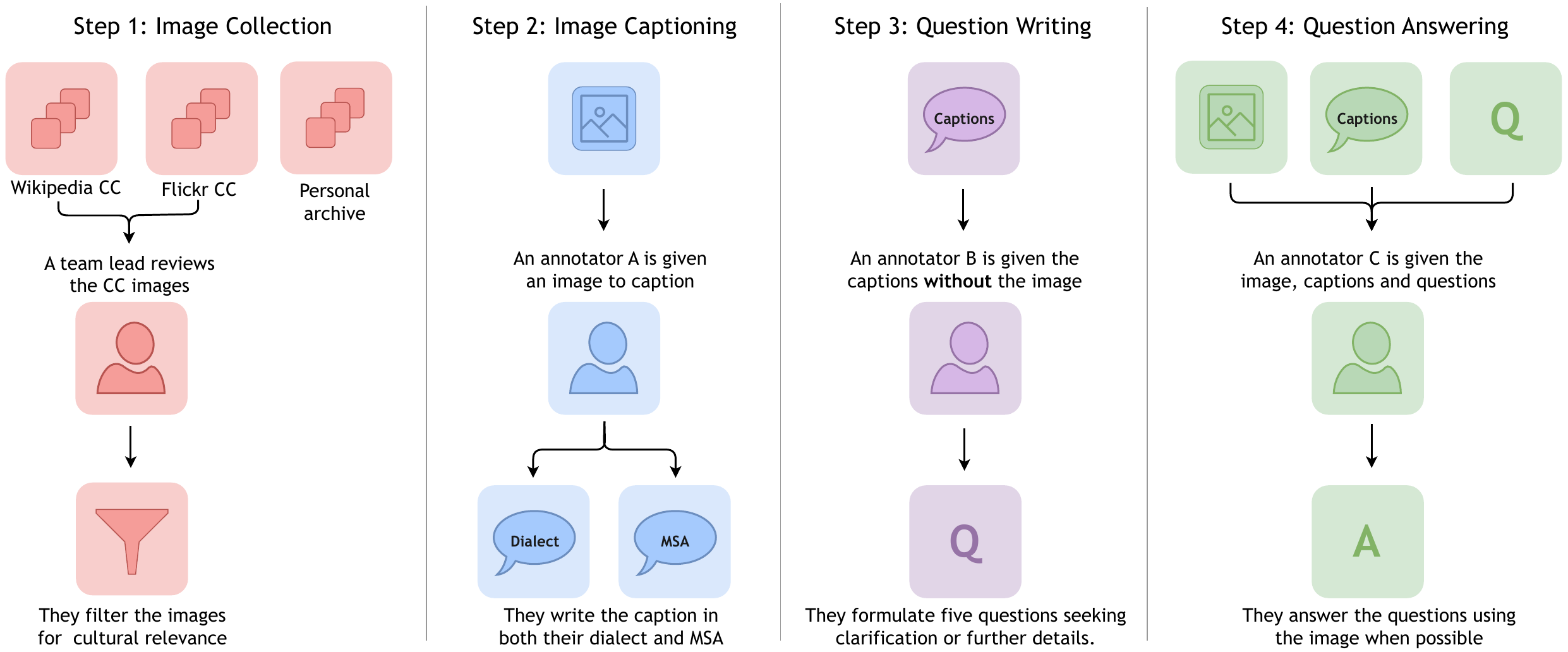}
    \caption{Data collection pipeline. 
    }
    \label{fig:annot-pipeline}
\end{figure*}

\subsection{Image Captioning}

You are presented with a photo that depicts a scene from daily life (e.g., food, clothing, homeware), social life (e.g., public transport, road signs, public ads), or urban objects from your area. Your task is to write a description of this photo in Arabic.

\subsubsection*{Steps for Writing}
\begin{enumerate}[nosep]
    \item \textbf{Analyze the photo:} Identify key elements, people, objects, actions, and any relevant background details.
    \item \textbf{Write the description of the photo:} The description should provide essential information. Typically, 15-25 words are sufficient. Describe everything that adds value and clarity.
    \item \textbf{Explain what is behind the scenes:} If necessary, describe the context of the photo using your background knowledge (e.g., where the photo could have been taken, whether the food in the photo is special, etc.).
    \item \textbf{Use everyday language:} Use ordinary informal language, but feel free to incorporate slang where appropriate.
\end{enumerate}

\subsubsection*{Hints for Creating a Better Description}
Your description should be detailed enough to give a clear idea of what is happening in the photo to someone who cannot see it. Try to include details that are specific to your culture or region. Here are some hints to help you:

\begin{itemize}[nosep]
    \item \textbf{Describe people, animals, objects, and key elements:} How they look and how they relate to each other in the physical space.
    \item \textbf{Describe interactions:} Who or what interacts with whom or what, and how they interact.
    \item \textbf{Include implicit details:} Add information that is not explicitly presented in the photo if it helps convey the image better. For example, if you can tell from people’s attire that this is a wedding party, even though there is no visible banner stating so, mention it in the description.
    \item \textbf{Use precise terminology:} ``Cat'' is better than ``animal'', and ``Siamese cat'' is better than ``cat.''
    \item \textbf{Rely on everyday knowledge and culture, but avoid over-fantasizing:} You do not need to create a story or a plot, but you should be as precise as possible in your description.
\end{itemize}

\subsection{Question Writing}

You are presented with a description of a photo, but you do not have access to the photo itself. Your task is to ask five questions that will help you better understand what is happening in the photo and refine the description.

\subsubsection*{Steps for Writing}
\begin{enumerate}[nosep]
    \item \textbf{Carefully read the description:} Identify parts that are unclear, ambiguous, or lacking in detail.
    \item \textbf{Formulate a question:} Craft a question to clarify ambiguities or add relevant details to the photo description.
    \item \textbf{Use everyday language:} Use ordinary, informal language, but feel free to incorporate slang where appropriate.
\end{enumerate}

\subsubsection*{Hints for Creating Better Questions}
\begin{itemize}[nosep]
    \item \textbf{Pay attention:} Do not ask for details that are already provided. For example, if the description states, “The photo shows a woman in a red dress,” you should not ask, “What color is the dress in the photo?”
    \item \textbf{Keep questions concise:} Questions should be no longer than one sentence. There is no need to provide additional context within the question.
    \item \textbf{Base your questions on the description:} You can inquire about people, animals, or objects mentioned—how they look, what people are wearing, what they are doing, how they relate to each other in physical space, and how they interact.
    \item \textbf{Ask about background details:} Consider why people are dressed a certain way, why they are performing specific actions, or why certain objects are present.
    \item \textbf{Inquire about future events:} Ask what might happen next—what people will do right after the described scene, or what will happen to the objects mentioned.
    \item \textbf{Request emotional or aesthetic judgments:} Ask whether the photo looks nice, whether it would work as a postcard, or whether it would make a good wall print.
    \item \textbf{Avoid unnecessary repetition:} You do not need to repeat the exact wording from the description in your question. For example, if the description states, “The picture shows an empty street with a single car passing by,” you do not have to use the word “car” in your question. Instead of asking, “What color is the car?” you can simply ask, “What color is it?”
\end{itemize}

\subsection{Question Answering}

You are presented with a photo that shows a scene from daily life (e.g., food, clothing, homeware), social life (e.g., public transport, road signs, public ads), or urban objects from your area, along with a description of this photo and five questions asking to clarify missing information from the photo. Your task is to answer the questions.

\subsubsection*{Steps for Writing}
\begin{enumerate}[nosep]
    \item \textbf{Analyze the photo:} Identify key elements, people, objects, actions, and any relevant background details.
    \item \textbf{Carefully read the description and the questions:} Identify what is unclear and missing in the description.
    \item \textbf{Answer the questions:} Provide a clear and detailed answer based on the photo to clarify or add to its description. Aim for 2-3 sentences.
    \item \textbf{Use everyday language:} Use ordinary, informal language, but feel free to use slang words where necessary.
    \item \textbf{Revise, Edit, Submit.}
\end{enumerate}

\subsubsection*{Hints for Creating Better Answers}
\begin{itemize}[nosep]
    \item \textbf{Take your time to carefully look over the photo:} Pay attention even to the smallest details before answering each question.
    \item \textbf{Base your answer on the photo or your cultural knowledge:} You do not need to create a story or explanation if it cannot be gathered from the photo.
    \item \textbf{If the question cannot be answered:} If something is not clear from the photo or your cultural knowledge, choose the option \texttt{<Cannot tell from the picture>}.
    \item \textbf{If details are already mentioned in the description:} You may simply copy the answer from there if the question asks for details that have already been stated.
\end{itemize}

\subsection{Human Evaluation of Image Captioning} 

You are presented with an image and an image caption — a short text that describes the content of the image. You need to look closely at the image, read its caption and evaluate the caption according to the following five criteria.
Evaluate each criterion on a scale from 1 to 5, where 1 means very bad, 3 means neutral, and 5 means excellent.
Be lenient; when in doubt, don’t be afraid to give a high score.

\noindent \textbf{Consistency:} Does the caption match what is actually shown in the image? It should avoid adding details that are not visible.

\noindent \textbf{Relevance:} Does the caption mention the most important elements in the image? It should focus on the main subjects without omitting key details.

\noindent \textbf{Fluency:} Evaluate how naturally and smoothly the text reads. Consider clarity, word choice, and overall ease of understanding. A fluent text should be easy to read, free of language errors, and sound natural. 

\noindent \textbf{Dialect authenticity:} How well does the caption represent the spoken dialect in your country? Does it use words and phrases that people in your country commonly would use?

\noindent Note that fluency and dialectal language are not the same. Fluency evaluates how natural and correct the caption is, while dialectal language assesses how dialectal or regional the caption sounds. A caption might be non-fluent but still dialectal or attempting to sound dialectal.

%% file: eval_prompts.tex
\begin{figure}[h]
    \centering
    \tcbset{colframe=black, colback=gray!10, arc=5mm}
    \begin{tcolorbox}
    \small
    \textbf{You are an expert evaluator assessing the quality of an Arabic image caption.} \textbf{You will be given an image, a reference caption, and a caption to evaluate}. Your task is to carefully analyze all three and evaluate the given caption based on four criteria: \textbf{Consistency, Relevance, Fluency, and Dialect Authenticity.}

    Evaluate each criterion on a scale from 1 to 5, where 1 means very bad, 3 means neutral, and 5 means excellent.

    \textbf{Consistency:} Does the caption match what is actually shown in the image? It should avoid adding details that are not visible.

    \textbf{Relevance:} Does the caption mention the most important elements in the image? It should focus on the main subjects without omitting key details.

    \textbf{Fluency:} Evaluate how naturally and smoothly the text reads. Consider clarity, word choice, and overall ease of understanding. A fluent text should be easy to read, free of language errors, and sound natural.

    \textbf{Dialect Authenticity:} How well does the caption represent the spoken dialect in \texttt{\{country\}}? Does it use words and phrases that people in this country commonly would use?

    \textbf{Reference Caption:} \texttt{\{reference\}} \\
    \textbf{Generated Caption:} \texttt{\{generated\}}

    \textbf{Output Format (do not add any additional information):} \\
    Consistency: X/5 \\
    Relevance: X/5 \\
    Fluency: X/5 \\
    Dialect Authenticity: X/5
    \end{tcolorbox}
    \caption{Evaluation prompt for assessing Arabic image captions \textbf{using both the image and the reference caption}. The evaluation is based on four criteria: Consistency, Relevance, Fluency, and Dialect Authenticity, following a structured format and a five-point rating scale. For MSA, only the first three criteria (Consistency, Relevance, and Fluency) were included in the prompt, while Dialect Authenticity was omitted.}
\end{figure}

\begin{figure}[h]
    \centering
    \tcbset{colframe=black, colback=gray!10, arc=5mm}
    \begin{tcolorbox}
    \small
    \textbf{You are an expert evaluator assessing the quality of an Arabic image caption.} \textbf{You will be given a reference caption and a caption to evaluate}. Your task is to carefully compare the evaluated caption to the reference caption and assess it based on four criteria: \textbf{Consistency, Relevance, Fluency, and Dialect Authenticity.}

    Evaluate each criterion on a scale from 1 to 5, where 1 means very bad, 3 means neutral, and 5 means excellent.

    \textbf{Consistency:} Does the evaluated caption match the reference caption in meaning and key details? It should avoid adding information that is not present in the reference caption or contradicting its content.

    \textbf{Relevance:} Does the evaluated caption mention the most important elements described in the reference caption? It should focus on the main subjects without omitting key details.

    \textbf{Fluency:} Evaluate how naturally and smoothly the text reads. Consider clarity, word choice, and overall ease of understanding. A fluent text should be easy to read, free of language errors, and sound natural.

    \textbf{Dialect Authenticity:} Check how well the caption represents the dialect spoken in \texttt{\{country\}}. Does it use words and phrases that people in this country commonly use?

    \textbf{Reference Caption:} \texttt{\{reference\}} \\
    \textbf{Generated Caption:} \texttt{\{generated\}}

    \textbf{Output Format (do not add any additional information):} \\
    Consistency: X/5 \\
    Relevance: X/5 \\
    Fluency: X/5 \\
    Dialect Authenticity: X/5
    \end{tcolorbox}
    \caption{Evaluation prompt for assessing Arabic image captions \textbf{using only the reference caption}. The evaluation is based on four criteria: Consistency, Relevance, Fluency, and Dialect Authenticity, following a structured format and a five-point rating scale. For MSA, only the first three criteria (Consistency, Relevance, and Fluency) were included in the prompt, while Dialect Authenticity was omitted.}
\end{figure}

\begin{figure}[h]
    \centering
    \tcbset{colframe=black, colback=gray!10, arc=5mm}
    \begin{tcolorbox}
    \small
    \textbf{You are an expert evaluator assessing the quality of an answer to a question in dialectal Arabic.} You will be given an image, a question, a reference answer, and an answer to evaluate. Your task is to carefully analyze all four and evaluate the given answer based on four criteria: \textbf{Consistency, Relevance, Fluency, and Dialect Authenticity.}

    Evaluate each criterion on a scale from 1 to 5, where 1 means very bad, 3 means neutral, and 5 means excellent.

    \textbf{Consistency:} Does the answer correctly address the question while accurately describing the content of the image? It should avoid adding details that are not visible.

    \textbf{Relevance:} Does the answer provide the most important details necessary to respond to the question based on the image? It should focus on the main subjects without omitting key details.

    \textbf{Fluency:} Evaluate how naturally and smoothly the answer reads. Consider clarity, word choice, and overall ease of understanding. A fluent answer should be easy to read, free of language errors, and sound natural.

    \textbf{Dialect Authenticity:} How well does the answer represent the spoken dialect in \texttt{\{country\}}? Does it use words and phrases that people in this country commonly would use?

    \textbf{Question:} \texttt{\{question\}} \\
    \textbf{Reference Answer:} \texttt{\{reference\}} \\
    \textbf{Generated Answer:} \texttt{\{generated\}}

    \textbf{Output Format (do not add any additional information and do not add explanations at the end of the evaluation):} \\
    Consistency: X/5 \\
    Relevance: X/5 \\
    Fluency: X/5 \\
    Dialect Authenticity: X/5
    \end{tcolorbox}
    \caption{Evaluation prompt for assessing answers to questions in dialectal Arabic using an image, a question, and a reference answer. The evaluation is based on four criteria: Consistency, Relevance, Fluency, and Dialect Authenticity, following a structured format and a five-point rating scale.}
\end{figure}

\begin{figure}[h]
    \centering
    \tcbset{colframe=black, colback=gray!10, arc=5mm}
    \begin{tcolorbox}
    \small
    \textbf{You are an expert evaluator assessing the type of a question in dialectal Arabic.} You will be given a question and must determine its type based on the following classification:

    \textbf{Classification:}
    \begin{itemize}[left=0pt, labelsep=10pt, itemsep=1.5ex, label={}]
        \item \begin{arabtext}
            \textbf{وصف:} أسئلة تهدف إلى الحصول على تفاصيل أو شرح عن موضوع أو حالة معينة.
        \end{arabtext}
        \item \begin{arabtext}
            \textbf{عد:} أسئلة تتعلق بعدد الأشياء أو الكميات أو تكرار حدوث شيء معين.
        \end{arabtext}
        \item \begin{arabtext}
            \textbf{تحقق:} أسئلة تهدف إلى التحقق من صحة أو خطأ معلومة أو حقيقة معينة.
        \end{arabtext}
        \item \begin{arabtext}
            \textbf{تصنيفي:} أسئلة تهدف إلى تصنيف أو تقسيم شيء ما إلى مجموعات أو أنواع أو فئات محددة.
        \end{arabtext}
    \end{itemize}

    \textbf{Examples:}
    \begin{itemize}[left=0pt, labelsep=10pt, itemsep=1.5ex, label={}]
        \item \begin{arabtext}
            \textbf{سؤال:} شو لون عباية الحرمة اللي فالصورة؟ \\
            \textbf{النوع:} وصف
        \end{arabtext}
        \item \begin{arabtext}
            \textbf{سؤال:} الناس اللي على السقف لابسين إيه؟ \\
            \textbf{النوع:} وصف
        \end{arabtext}
        \item \begin{arabtext}
            \textbf{سؤال:} كم عدد البلدان التي تحدث فيها هذه الظاهرة؟ \\
            \textbf{النوع:} عد
        \end{arabtext}
        \item \begin{arabtext}
            \textbf{سؤال:} مبين شو اسم المطعم فالصورة؟ \\
            \textbf{النوع:} تحقق
        \end{arabtext}
        \item \begin{arabtext}
            \textbf{سؤال:} هل اليهال من نفس الأعمار؟ \\
            \textbf{النوع:} تحقق
        \end{arabtext}
        \item \begin{arabtext}
            \textbf{سؤال:} الشجر الي فالصورة شجر زينة ولا شجر مثمر؟ \\
            \textbf{النوع:} تصنيفي
        \end{arabtext}
        \item \begin{arabtext}
            \textbf{سؤال:} هاد الناس اللي فكوزينة واش رجال ولا عيالات؟ \\
            \textbf{النوع:} تصنيفي
        \end{arabtext}
    \end{itemize}

    \textbf{Task:}
    \begin{itemize}[left=0pt, labelsep=10pt, itemsep=1.5ex, label={}]
        \item \begin{arabtext}
            \textbf{السؤال:} 
         \end{arabtext}
    
            \texttt{\{question\}} \\
        \begin{arabtext}
            \textbf{النوع:} 
        \end{arabtext}    
            \texttt{\{type\}}
        
    \end{itemize}
    \end{tcolorbox}
    \caption{Question Type Identification Prompt. The task is to determine the type of a given question in dialectal Arabic based on a predefined classification.}
    \label{fig:question_type_prompt}
\end{figure}

\begin{figure}[h]
    \centering
    \tcbset{colframe=black, colback=gray!10, arc=5mm}
    \begin{tcolorbox}
    \small
    \textbf{You are an expert evaluator identifying the main topic of a description in Arabic.} You will be given a description and must determine its main topic based on the following process:

    \textbf{Process:}
    \begin{itemize}[left=0pt, labelsep=10pt, itemsep=1.5ex, label={}]
        \item \begin{arabtext}
            \textbf{الخطوة الأولى:} قم بتعيين موضوع رئيسي لكل وصف باللغة العربية الفصحى.
        \end{arabtext}
        \item \begin{arabtext}
            \textbf{الخطوة الثانية:} قم بتجميع المواضيع المُحددة في فئات نهائية بناءً على التشابه بينها.
        \end{arabtext}
    \end{itemize}

    \textbf{Examples:}
    \begin{itemize}[left=0pt, labelsep=10pt, itemsep=1.5ex, label={}]
        \item \begin{arabtext}
            \textbf{الوصف:} صورة لنخلة تظهر من الأعلى، حيث يغطّيها الكثير من الأوراق الخضراء، وتوجد بها ثمار بلح أخضر صغير لم ينضج بعد. هناك بعض الخوص يخرج من جذع النخلة وأوراقها. تم التقاط الصورة من زاوية منخفضة وقريبة من قمة النخلة. \\
            \textbf{الموضوع:} طبيعة
        \end{arabtext}
        \item \begin{arabtext}
            \textbf{الوصف:} صورة لساحة مكان تشبه المول التجاري. الساحة مغطاة بأرضية سيراميك كبيرة بلون متدرج من البيج إلى البني. المكان عبارة عن مبنى مكون من طابق واحد، لونه بين البيج والرمادي، وجميع نوافذه من الزجاج. توجد في المبنى فواصل طويلة باللون الأحمر والأزرق والأصفر. في الساحة جزء مغطى بالزرع، وأغلب المساحة مظللة بشيء يشبه التندة. كما يوجد مجسم يشبه الساقية، بالإضافة إلى وجود أشخاص وأطفال في المكان. \\
            \textbf{الموضوع:} أماكن
        \end{arabtext}
    \end{itemize}

    \textbf{Task:}
    \begin{itemize}[left=0pt, labelsep=10pt, itemsep=1.5ex, label={}]
        \item \begin{arabtext}
            \textbf{الوصف:} 
        \end{arabtext}
            \texttt{\{caption\}} \\
        \item \begin{arabtext}
            \textbf{الموضوع:} 
        \end{arabtext}
            \texttt{\{topic\}}
    \end{itemize}
    \end{tcolorbox}
    \caption{Topic Identification Prompt. The task is to determine the main topic of a given caption in Arabic.}
    \label{fig:topic_identification_prompt}
\end{figure}

%% file: app_topics_grouping.tex
\begin{table}[h!]
\centering
\tiny
\renewcommand{\arraystretch}{1}
\begin{tabular}{l p{12cm}}
\toprule
\vspace{2pt}

\textbf{Places} & { \begin{arabtext} سوق, مسجد, مدينة, الصحراء,  قرية, حديقة, كورنيش, قرية جبلية, مدينة ساحلية, منطقة سكنية, منطقة جبلية,     مطار, ميناء, مركز تجاري, مركز الخدمات الطبية, مكتبة الجامعة, مقبرة, مبنى حكومي, مبنى البريد, محطة وقود\end{arabtext}} \\
\textbf{Translation} & Market, Mosque, City,  Desert, Village, Park, Corniche, Mountain Village, Coastal City, Residential Area, Mountain Area,   Airport, Port, Mall, Medical Center, University Library, Cemetery, Government Building, Post Office, Gas Station \\
\midrule
\textbf{Celebrations} & \begin{arabtext} عيد الميلاد, احتفال, حفل زفاف, حفل توزيع الجوائز, مهرجان, مهرجان فولكلوري, اجتماع, ندوة, محاضرة, عرض موسيقي, عرض تقديمي, تظاهرة, حملة توعية, حملة تطوعية \end{arabtext} \\
\textbf{Translation} & Christmas, Celebration, Wedding, Award Ceremony, Festival, Folk Festival, Meeting, Seminar, Lecture, Music Performance, Presentation, Demonstration, Awareness Campaign, Volunteer Campaign \\
\midrule
\textbf{Arts and Culture} & \begin{arabtext}  رقص تقليدي, عرض موسيقي, فيلم, حفلة, موسيقية, ورشة عمل فنية, تحضير الطعام التقليدي, تحضير الحلويات, الطبخ التقليدي, الفن التقليدي, التراث, الأسواق التقليدية, الحرف التقليدية, منتجات خزفية, منتوجات تقليدية, تذكارات  نقش الحناء, نقش البلاط,  تلوين المزهريات \end{arabtext} \\
\textbf{Translation} & Traditional Dance, Music Performance, Film, Party, Musical, Art Workshop, Traditional Food Preparation, Sweets Preparation, Traditional Cooking, Traditional Art, Heritage, Traditional Markets, Traditional Crafts, Ceramic Products, Traditional Products, Souvenirs, Henna Art, Tile Art, Pottery Decoration \\
\midrule
\textbf{Nature} &  \begin{arabtext} غابة, طبيعة, وادي رم, وادي دادس, شجرة نخيل, غروب الشمس, منظر طبيعي, واحة صحراوية, جبال, الصحراء, الطبيعة المغربية, الشتاء, الربيع, الزقاق المغربي, الحديقة, حديقة عامة, حديقة حيوانات, حديقة ماجوريل, الطبيعة المغربية \end{arabtext}  \\
\textbf{Translation} & Forest, Nature, Wadi Rum, Dades Valley, Palm Tree, Sunset, Scenic View, Desert Oasis, Mountains, Desert, Moroccan Nature, Winter, Spring, Moroccan Alley, Garden, Public Park, Zoo, Majorelle Garden, Moroccan Nature \\
\midrule
\textbf{Education} & \begin{arabtext} مدرسة, الثانوية التأهيلية الحسن الثاني, تعليم اللغة الفرنسية, تدريب المعلمين, محل خياطة,  مخيم كشافة, ورشة عمل, مكتبة الجامعة, المدرسة, فصل دراسي \end{arabtext}\\
\textbf{Translation} & School, Hassan II High School, French Language Education, Teacher Training, Tailor Shop,  Scout Camp, Workshop, University Library, School, Classroom \\
\midrule
\textbf{Transport} & \begin{arabtext} محطة حافلات, محطة ترامواي, محطة سيارات الأجرة, سيارة أجرة, طريق, طريق سيار, طريق نائية, سيارات الأجرة, الطاكسيات, ترامواي, قطار, القطارات, موقف سيارات, موقف سيارات الأجرة, شاحنة, سيارة قديمة, توك توك, حافلة \end{arabtext} \\
\textbf{Translation} & Bus Station, Tram Station, Taxi Stand, Taxi, Road, Highway, Remote Road, Taxis, Taxis, Tram, Train, Trains, Parking Lot, Taxi Parking, Truck, Old Car, Tuk Tuk, Bus \\
\midrule
\textbf{Characters} & \begin{arabtext} الملك محمد السادس, الأميرة للا سلمى, شخصيات سياسية, مقدم, رجل مسن, نساء من شمال المغرب, رجل, فتيات, أطفال, عامل بناء, بائع, بائع متجول \end{arabtext} \\
\textbf{Translation} & King Mohammed VI, Princess Lalla Salma, Political Figures, Presenter, Elderly Man, Women from Northern Morocco, Man, Girls, Children, Construction Worker, Seller, Street Vendor \\
\midrule
\textbf{Food and Beverages} & \begin{arabtext} طعام, فلافل, عصير البرتقال, شاي, شاي مغربي, خبز, الكبسة, الكسكس, الحناء, مخللات, تحلية, عصائر طبيعية, وجبة طعام, فواكه, زيت الأركان, أملو, هريس, فواكه جافة, خبز مقلي, الكنافة, غزل الصوف, الخبز, الزقاق, الطعام التقليدي, المأكولات المغربية \end{arabtext} \\
\textbf{Translation} & Food, Falafel, Orange Juice, Tea, Moroccan Tea, Bread, Kabsa, Couscous, Henna, Pickles, Dessert, Fresh Juices, Meal, Fruits, Argan Oil, Amlou, Porridge, Dried Fruits, Fried Bread, Kunafa, Wool Spinning, Bread, Alley, Traditional Food, Moroccan Cuisine \\
\midrule
\textbf{Sports} & \begin{arabtext} رياضة, كرة القدم,  رياضات قتالية, فروسية, سباق السيارات, ركوب الجمل, المنتخب المغربي, المنتخب المغربي لكرة القدم, مباراة رياضية \end{arabtext} \\
\textbf{Translation} & Sports, Football, Martial Arts, Equestrianism, Car Racing, Camel Riding, Moroccan National Team, Moroccan National Football Team, Sports Match \\

\midrule
\textbf{Trade} & \begin{arabtext} التجارة, محل زيتون, محل مكسرات, محل بيع المنتجات التقليدية, محل بيع التمر, محل خياطة, محل بيع الفاكهة, محل الأعشاب والتوابل, محل بيع الفواكه, محل بيع الخبز, محل بيع الخضار, محل بيع المكسرات, محل بيع الحلوى \end{arabtext} \\
\textbf{Translation} & Commerce, Olive Shop, Nut Shop, Traditional Products Shop, Date Shop, Tailor Shop, Fruit Shop, Herbs and Spices Shop, Fruit Shop, Bread Shop, Vegetable Shop, Nut Shop, Sweet Shop \\
\midrule
\textbf{Technology} &  \begin{arabtext} معدات طبية, تسوق, عرض مشروع, عرض تقديمي, عرض جوي, تكنولوجيا, اتصالات المغرب, تكنولوجيا المعلومات \end{arabtext}  \\
\textbf{Translation} & Medical Equipment, Shopping, Project Presentation, Presentation, Aerial Show, Technology, Maroc Telecom, Information Technology \\
\midrule
\textbf{Games} & \begin{arabtext} ألعاب, لعبة الدومينو, ألعاب أطفال, ألعاب تقليدية, ألعاب رياضية \end{arabtext} \\
\textbf{Translation} & Games, Dominoes,  Children's Games, Traditional Games, Sports Games \\
\midrule
\textbf{Others} &   \begin{arabtext} الخطوط الملكية المغربية, الشرطة, الحرس الملكي, الجيش, الاستعمار الفرنسي في المغرب, دبلوماسية, البرلمان المغربي, المديرية العامة للأمن الوطني, الهيئة الوطنية للسلامة المرورية, الهيئة الوطنية للوقاية من حوادث السير \end{arabtext}\\
\textbf{Translation} & Royal Air Maroc, Police, Royal Guard, Army, French Colonization in Morocco, Diplomacy, Moroccan Parliament, Directorate General of National Security, National Road Safety Authority, National Traffic Accident Prevention Authority \\

\bottomrule

\end{tabular}
\caption{Topic Categories.}\label{tab:Topic Categories}
\end{table}